\documentclass[letterpaper, 10 pt, journal, twoside]{IEEEtran}

\usepackage{lipsum}  
\usepackage{amsmath, amsfonts}
\usepackage{algorithmic}
\usepackage{array}
\usepackage[caption=false,font=normalsize,labelfont=sf,textfont=sf]{subfig}
\usepackage{textcomp}
\usepackage{stfloats}
\usepackage{url}
\usepackage{verbatim}
\usepackage{graphicx}
\usepackage{balance}
\usepackage{hyperref}
\usepackage{cite}
\usepackage{bm}
\usepackage{hologo} 
\usepackage{booktabs} 
\usepackage{threeparttable} 
\usepackage{mathtools}

\hypersetup{
    colorlinks=false,
    citecolor=blue,
    linkcolor=red,
    filecolor=green,
    urlcolor=cyan,
    pdftitle={Learning Generalizable Policy for Obstacle-Aware Autonomous Drone Racing},
    pdfpagemode=FullScreen,
}

\title{Learning Generalizable Policy for Obstacle-Aware Autonomous Drone Racing}

\author{Yueqian Liu \thanks{This paper is part of the author's M.Sc. thesis supervised by Ir. Hang Yu and Dr. Ir. Christophe De Wagter, at MAVLab TU Delft. Full thesis is available at \texttt{\textmd{\url{https://repository.tudelft.nl}}}.}}

\begin{document}

\maketitle

\begin{abstract}
Autonomous drone racing has gained attention for its potential to push the boundaries of drone navigation technologies. While much of the existing research focuses on racing in obstacle-free environments, few studies have addressed the complexities of obstacle-aware racing, and approaches presented in these studies often suffer from overfitting, with learned policies generalizing poorly to new environments. This work addresses the challenge of developing a generalizable obstacle-aware drone racing policy using deep reinforcement learning. We propose applying domain randomization on racing tracks and obstacle configurations before every rollout, combined with parallel experience collection in randomized environments to achieve the goal. The proposed randomization strategy is shown to be effective through simulated experiments where drones reach speeds of up to 70 km/h, racing in unseen cluttered environments. This study serves as a stepping stone toward learning robust policies for obstacle-aware drone racing and general-purpose drone navigation in cluttered environments. Code is available at \texttt{\textmd{\url{https://github.com/ErcBunny/IsaacGymEnvs}}}.
\end{abstract}


\section{Introduction}

\IEEEPARstart{A}{utonomous} drone navigation has emerged as a critical area of research, driven by the growing demand for drones in industries such as delivery, inspection, and emergency response. Drone racing, with its emphasis on minimum-time navigation, has become a benchmark task for testing advanced autonomous systems aiming to navigate at high speeds while avoiding obstacles in partially or fully unknown environments. Drone racing originally began as a competitive sport where human pilots control agile drones via radio to fly through a racing track as fast as possible while avoiding potentially present obstacles. This requires precision, quick reflexes, and expert navigation skills. In autonomous drone racing, human pilots are replaced by algorithms and artificial intelligence (AI). This introduces the challenge for algorithms and AI of matching human-level performance and adaptability.

There have been several global autonomous drone racing events, including the 2016-2019 IROS Autonomous Drone Racing (ADR) competitions \cite{adr1617, adr18_Beauty_and_the_Beast}, the 2019 AlphaPilot Challenge \cite{alphapilot_rpg, alphapilot_mavlab}, the 2019 NeurIPS Game of Drones \cite{GameofDr}, and the 2022-2023 DJI RMUA UAV Challenges \cite{rmua22_uniminco, rmua23}. The tracks in early competitions, such as the IROS ADR, AlphaPilot, and Game of Drones, are situated in less cluttered spaces, allowing drones to complete the tracks without considering obstacle avoidance. However, for tracks in cluttered environments, such as those in the more recent DJI RMUA Challenges, the absence of obstacle awareness could cause crashing. Additionally, in human-piloted drone racing, such as the Drone Racing League competitions, and in drone racing video games, there are plenty of tracks that require obstacle avoidance.

Although autonomous drone racing has received significant attention, much of the research has been limited to obstacle-free scenarios \cite{adrsurvey_rpg}. Obstacle-free scenarios do not reflect the complexities encountered in real-world tasks where obstacle avoidance is necessary. Recognizing this, researchers have been exploring ways to integrate drone racing with obstacle avoidance through various approaches. Path-planning and optimization-based approaches can achieve short lap times \cite{time_opt_obs_sb} and strike a balance between lap times and computational efficiency \cite{trr, fastracing, rmua22_uniminco}, but rely on carefully designed algorithms and may experience performance degradation when model mismatches occur. Current learning-based approaches \cite{time_opt_obs_rl, time_opt_obs_rl_vis} leverage reinforcement learning (RL) and imitation learning (IL) to train neural policies capable of making low-latency decisions that result in aggressive and collision-free trajectories. However, these policies do not generalize well to new racing tracks or different obstacle configurations.

\begin{figure}[t]
    \centering
    \includegraphics[width=\linewidth]{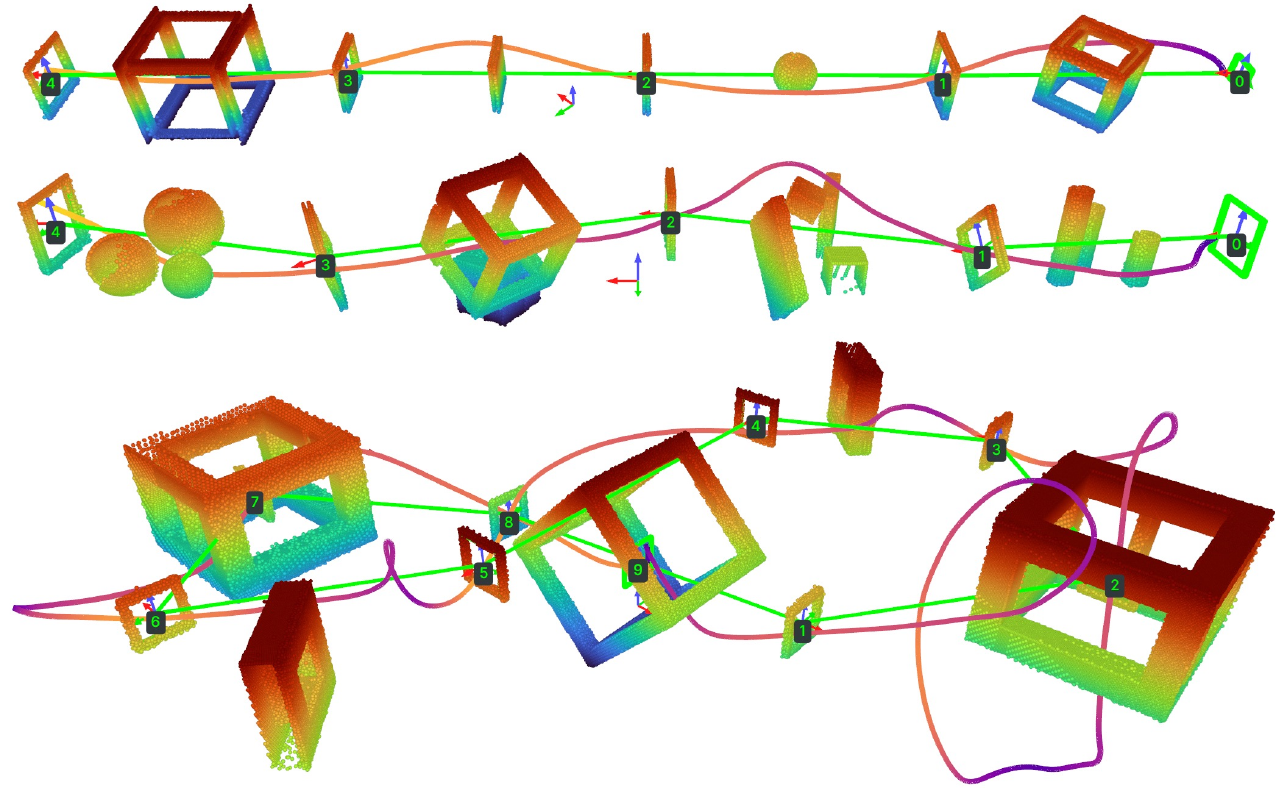}
    \caption{Trajectories of successful rollouts of a single policy on multiple different racing tracks with obstacles designed to block flight paths.}
    \label{fig:demo}
\end{figure}

This paper aims to enhance the generalization ability of learned policies. Specifically, the goal is to develop a single policy capable of navigating a quadcopter through various racing tracks with obstacles, without requiring additional tuning after training. Drawing inspiration from works on learning drone navigation in cluttered environments \cite{mavrl, ntnu-rlflight} and generalizable obstacle-free drone racing \cite{adrdrl}, which all involve training the policy in multiple randomized environments, we propose applying the same strategy, domain randomization \cite{domain_rand} over racing tracks, to expose the agent to a diverse set of environments. This allows the policy to learn the underlying navigation ``skills'' while not relying on unique observations associated with one or a few training environments. Simulated experiments verify that the resulting policy can indeed generalize to unseen racing tracks while avoiding obstacles in unseen sizes and shapes. Several successful examples are shown in Fig \ref{fig:demo}. In summary, the main contributions of this study are:

\begin{itemize}
    \item We verify the effectiveness of applying domain randomization to encourage the learning of generalizable skills.
    \item We present the first generalizable neural policy for the obstacle-aware drone racing task, where the policy directly maps observations to low-level commands.
    \item We open-source tools and reusable modules to facilitate research and development in both obstacle-free and obstacle-aware drone racing.
\end{itemize}

\section{Related Work}

\subsection{Obstacle-Free Autonomous Drone Racing}

Optimization-based methods have been widely applied to the task of drone racing. For static obstacle-free racing tracks, time-optimal trajectories passing through gates' center waypoints can be generated using optimization with Complementary Progress Constraint (CPC) \cite{timeopt_wp_cpc}. However, this approach is computationally expensive and struggles to adapt to changing track layouts. To reduce computational overhead, Model Predictive Contouring Control (MPCC) is introduced \cite{mpcc}. MPCC has been further extended to include an online reference path generation module to adapt to dynamic tracks and handle external disturbances \cite{mpcc_online}. A more recent study \cite{timopt_wayspace} demonstrates that lap times can be further reduced by exploiting the spatial potential of the gates.

Reinforcement Learning has also emerged as a promising approach for autonomous drone racing. Near-time-optimal agile flight can be achieved through state-based RL \cite{adrdrl}. Although the learned policy results in slightly longer lap times than the CPC method, it handles variations in gate poses and generalizes to unseen tracks. Furthermore, state-based policies can serve as teacher policies within the IL framework, enabling the training of purely vision-based student policies \cite{visual_drl}. In a follow-up study \cite{bootstrap}, the student policy is further fine-tuned using RL. These two studies show the feasibility of high-speed agile flight using AI with the same input-output modalities as human pilots.

Reinforcement learning based approaches offer several advantages over optimization-based methods. These include improved lap times and higher success rates during real-world flights, where unmodeled effects and disturbances are non-negligible \cite{adr_oc_vs_rl}. However, deploying policies in real-world scenarios is challenging, requiring closing the sim-to-real gap and careful system engineering. The Swift system \cite{swift} demonstrates that, by bridging the gap via fine-tuning using data-driven residual models, AI systems powered by RL policies can achieve performance on par with human champions.

\subsection{Obstacle-Aware Autonomous Drone Racing}

For the task of obstacle-Aware autonomous drone racing, leveraging optimization and planning, several methods have proven effective. The teach-and-repeat framework is widely used in autonomous robot missions, and has been applied to drone racing \cite{trr}. This framework enables the drone to fly through the track while avoiding previously unseen and dynamic obstacles. In static environments, Fast-Racing \cite{fastracing} provides a polynomial trajectory baseline for obstacle-aware autonomous drone racing. The winning solution of the 2022 DJI RMUA UAV Challenge \cite{rmua22_uniminco} also follows a polynomial-based trajectory but incorporates an additional online re-planning module to avoid dynamic obstacles and pass through moving gates. Moreover, Penick et al. \cite{time_opt_obs_sb} offer a sampling-based baseline aimed at finding time-optimal trajectories in cluttered environments, though this method struggles to scale with increasing environment complexity.

While RL-based methods have shown great promise in obstacle-free autonomous drone racing, achieving better lap times, disturbance rejection, and less compute latency, their application to obstacle-aware racing remains relatively sparse. To the best of our knowledge, only two studies \cite{time_opt_obs_rl, time_opt_obs_rl_vis} have addressed this challenging task. Furthermore, they all focus on completing a single predefined racing track in minimum time, not considering generalizing to unseen scenarios. Realizing this gap, we aim to explore methods that enhance policy generalization in obstacle-aware drone racing.

\subsection{Vision-Involved Navigation via Deep RL}

In the aforementioned works addressing obstacle-aware drone racing with RL, the policies fail to generalize to environments different from the ones they were trained on. To overcome this limitation, research on learning general-purpose navigation offers valuable insights and guidance.

Near-perfect discrete-action indoor navigation for ground robots has been demonstrated with DD-PPO \cite{ddppo}, in which training occurs across multiple indoor scenes to enhance generalization. The agent utilizes a policy network comprising a Convolutional Neural Network (CNN) as the encoder and a Long Short-Term Memory (LSTM) network. At first, this network is optimized as a whole using RL, which is inefficient. Follow-up works \cite{aux-pointgoalnav, aux-ddppo} show that using auxiliary tasks, such as predicting depth, inverse dynamics, and remaining distance to target, results in quicker policy convergence and better overall performance.

Besides using auxiliary tasks, modular learning is another approach to achieving efficient learning for vision-involved navigation tasks. Hoeller et al. \cite{quadrupednav} propose a modular learning framework for training a quadruped robot to navigate in cluttered dynamic environments. Here network modules are learned separately: once an upstream module is learned, it is frozen while downstream modules are optimized. MAVRL \cite{mavrl} and Kulkarni et al. \cite{ntnu-rlflight} also adopt a similar framework for drone navigation in clutter. All these methods utilize randomization of the training environments to promote generalization, which directly inspires our core idea for learning a generalizable policy in obstacle-aware drone racing.

\section{Methodology}

\subsection{Drone Model and Waypoints}
\label{subsec: drone_model_wp}

The drone is modeled as a rigid body with mass $m$ and moment of inertia $\bm{J}$. With the body frame attached to the center of gravity, the equations of dynamics can be written as:

\begin{equation}
\label{eq:rigid_body_dyn}
    \begin{bmatrix}
        \dot{\bm{p}}_{\mathcal{W}} \\
        \dot{\bm{q}}_{\mathcal{WB}} \\
        \dot{\bm{v}}_{\mathcal{W}} \\
        \dot{\bm{\omega}}_{\mathcal{B}} \\
        \dot{\bm{\Omega}} \\
    \end{bmatrix} = \begin{bmatrix}
        \bm{v}_{\mathcal{W}} \\
        \frac{1}{2}\bm{q}_{\mathcal{WB}} \otimes \begin{bmatrix}
                0 & \bm{\omega}_{\mathcal{B}}^{\mathsf{T}}
            \end{bmatrix}^{\mathsf{T}} \\
        \bm{g}_{\mathcal{W}} + \frac{1}{m} \bm{R}_{\mathcal{WB}} ( \bm{f}_{a} + \bm{f}_{d} ) \\
        \bm{J}^{-1} (\bm{\tau}_{a} + \bm{\tau}_{d} - \bm{\omega}_{\mathcal{B}} \times \bm{J}\bm{\omega}_{\mathcal{B}}) \\
        k_r (\bm{\Omega}_{s} - \bm{\Omega})
    \end{bmatrix} \text{,}
\end{equation}

\noindent where $\mathcal{W}$, $\mathcal{B}$ denote the world frame and the drone body frame, and $\bm{p}_{\mathcal{W}}$, $\bm{q}_{\mathcal{WB}}$, $\bm{v}_{\mathcal{W}}$, $\bm{\omega}_{\mathcal{B}}$, $\bm{g}_{\mathcal{W}}$ represent drone position, attitude quaternion, linear velocity, angular velocity, and gravitational acceleration, respectively. Rotation matrix of drone attitude is $\bm{R}_{\mathcal{WB}}$. Actuator wrench, or ``force and torque'', $(\bm{f}_a, \bm{\tau}_a)$ and aerodynamic drag wrench $(\bm{f}_d, \bm{\tau}_d)$ make up the total wrench acting on the rigid body. Rotor spinning dynamics is considered as a first-order lag model: the derivative of rotor angular velocities in rounds per second (RPM) $\dot{\bm{\Omega}}$ is the product of the rotor constant $k_r$ and the difference between steady-state RPM $\bm{\Omega}_s$ and current RPM $\bm{\Omega}$.

From rotor angular velocities and accelerations we can calculate the actuator wrenches:

\begin{equation}
\label{eq:actuator_wrench}
    \begin{bmatrix}
        \bm{f}_a \\
        \bm{\tau}_a
    \end{bmatrix}
    = 
    \begin{bmatrix}
        \sum_{i} \bm{f}_p(\bm{\Omega}_i) \\
        \sum_{i} \left(\bm{\tau}_p(\bm{\Omega}_i) + \bm{r}_i \times \bm{f}_p(\bm{\Omega}_i) + \zeta_{i} \bm{J}_r \dot{\bm{\Omega}}_{i}\right)
    \end{bmatrix} \text{,}
\end{equation}

\noindent where $\bm{f}_p(\bm{\Omega}_i)$ and $\bm{\tau}_p(\bm{\Omega}_i)$ represent the thrust force and torque generated by the $i$-th spinning propeller, respectively. They are calculated using polynomial models derived from the motor manufacturer's data. Here, $\bm{r}_i$ is the displacement of the $i$-th rotor relative to the drone's body frame origin, $\bm{J}_r$ denotes the moment of inertia of the rotor, which includes the propeller and the spinning parts of the motor, and $\zeta_i$ indicates the rotational direction of the $i$-th rotor.

The aerodynamic drag depends on the linear velocity, $\bm{v}_{\mathcal{B}} = \bm{R}_{\mathcal{WB}}^{-1} \bm{v}_{\mathcal{W}}$, and the angular velocity, $\bm{\omega}_{\mathcal{B}}$, in the body frame:

\begin{equation}
\label{eq:drag_wrench}
    \begin{bmatrix}
        \bm{f}_d \\
        \bm{\tau}_d
    \end{bmatrix}
    =
    -\frac{1}{2} \rho
    \begin{bmatrix}
        A_t \left(\bm{C}_0 \bm{v}_{\mathcal{B}}^2 + \bm{C}_1 \bm{v}_{\mathcal{B}}\right) \\
        A_r \left(\bm{C}_2 \bm{\omega}_{\mathcal{B}}^2 + \bm{C}_3 \bm{\omega}_{\mathcal{B}}\right)
    \end{bmatrix} \text{,}
\end{equation}

\noindent where $\rho$ is the air density, $A_t$ and $A_r$ are the effective areas responsible for generating translational and rotational drag, respectively. Coefficients $\bm{C}_i$ for $i$ from 0 to 3 are adjustable coefficients in the polynomial drag model.

The steady-state angular velocity is determined by the motor commands $\bm{u}_m$ using a polynomial model fitted to the motor manufacturer's data. Simulating how human operators control racing drones, we employ an angular velocity controller to translate control commands $\bm{a} \in [-1, 1]^4$ to motor commands $\bm{u}_m \in [0, 1]$ of the drone. Vector $\bm{a}$ contains 3 channels for body rates and 1 channel for the throttle, or collective thrust level. The angular velocity controller is derived from Betaflight \cite{betaflight} and is responsible for mapping control commands to desired angular velocity, running closed-loop control, and allocating the control to motor commands.

Apart from dynamics and control, the rigid body's collision geometry is coarsely approximated by the minimum body-frame axis-aligned bounding box of the real geometry. Regarding sensors, we assume we have access to accurate vehicle states including $\bm{p}_{\mathcal{W}}$, $\bm{q}_{\mathcal{WB}}$, $\bm{v}_{\mathcal{W}}$, and $\bm{\omega}_{\mathcal{B}}$. Furthermore, we attach a tilted depth camera to the front of the drone and set the depth sensing range to a relatively large value (e.g. 20 m) to mimic how human operators or depth estimation networks retrieve depth from monocular images.

Drone racing requires the drone to pass through gates from the correct side to the other. In the obstacle-aware scenario, there are possible additional requirements for the drone to avoid certain obstacles by following desired courses where no physical gates exist. We propose to model physical gates and course constraints as waypoints. A waypoint is defined as a finite-size rectangle plane parameterized by position $\bm{p}_{\text{wp}}$, attitude quaternion $\bm{q}_{\text{wp}}$, width and height of the valid pass-through region $(w_{\text{wp}}, h_{\text{wp}})$, and a binary parameter $g_{\text{wp}}$, which indicates the presence of physical bars around the waypoint. The waypoint frame, denoted by $\mathcal{G}$, is attached to the center of the rectangle, with the $x$-axis perpendicular to the plane, pointing towards the valid pass-through direction, and $y$-axis $z$-axis perpendicular to the sides of the rectangle. An illustration of waypoints can be found in Figure \ref{fig:wp_obs}(a).

\subsection{Task Formulation}

We formulate obstacle-aware autonomous drone racing as a partially observable Markov decision process (POMDP) \cite{pomdp}. POMDP models the agent-environment interaction and internal dynamics of the environment. The agent is an AI system that decides on the action to take based on observations from the environment. The environment is everything else that takes the agent's actions, updates environment states, computes rewards, and finally outputs the observations, closing the interaction loop. Actions are taken based on the policy $\pi$ that maps observations to actions, with the transition model of states, a trajectory $\bm{\tau}$ can be produced. The goal is to find the policy that maximizes the expectation of total discounted reward:

\begin{equation}
\label{eq:pomdp_goal}
    \underset{\pi}{\max}\  \mathbb{E}_{\bm{\tau}}\left[\sum_{t=0}^{\infty}\gamma^t r(t)\right] \text{,}
\end{equation}

\noindent where $\gamma \in [0,1)$ stands for the discount factor, and $r(t)$ denotes the reward as a function of time $t$.

\subsubsection{States}

States include every piece of necessary information to define the environment configuration. This may include drone rigid body states, camera transform, internal states of the actuator model and controllers, gate poses and sizes, obstacle shapes, and poses, etc.

\subsubsection{Action}

As discussed in the previous drone model section, we use the control commands denoted by $\bm{a}$ as the action to simulate human operators' control commands to a radio-controlled racing drone.

\subsubsection{Transition}

The transition of states of our environment is deterministic and only updates states associated with the drone model. Action $\bm{a}$ is turned into actuator wrenches through the angular velocity controller and Equation (\ref{eq:actuator_wrench}). Together with the drag wrenches in Equation (\ref{eq:drag_wrench}), drone states $\bm{p}_{\mathcal{W}}$, $\bm{q}_{\mathcal{WB}}$, $\bm{v}_{\mathcal{W}}$, and $\bm{\omega}_{\mathcal{B}}$ can be updated using Equation (\ref{eq:rigid_body_dyn}) with an integrator or a physics engine. States of obstacles and gates are fixed within an episode of the process.

\subsubsection{Reward Function}

The reward function is a weighted sum of reward terms including the progress reward $r_{\text{prog}}$, perception reward $r_{\text{prec}}$, command reward $r_{\text{cmd}}$, collision reward $r_{\text{col}}$, guidance reward $r_{\text{guid}}$, waypoint passing reward $r_{\text{wp}}$, timeout reward $r_{\text{time}}$, and finally the linear velocity reward $r_{\text{vel}}$. By representing weights collectively as a vector $\bm{\lambda}$, we can write the reward as:

\begin{equation}
    r = \begin{bmatrix}
        r_{\text{prog}} &
        r_{\text{prec}} &
        r_{\text{cmd}} &
        r_{\text{col}} &
        r_{\text{guid}} &
        r_{\text{wp}} &
        r_{\text{time}} &
        r_{\text{vel}}
    \end{bmatrix}\bm{\lambda} \text{.}
\end{equation}

\noindent Terms $r_{\text{prog}}$, $r_{\text{prec}}$, $r_{\text{cmd}}$, and $r_{\text{col}}$ are formulated as in Swift \cite{swift}. The guidance reward is extended based on the safety reward seen in \cite{adrdrl}. The remaining ones are additional terms proposed in this study.

\begin{figure}[t]
    \centering
    \includegraphics[width=\linewidth]{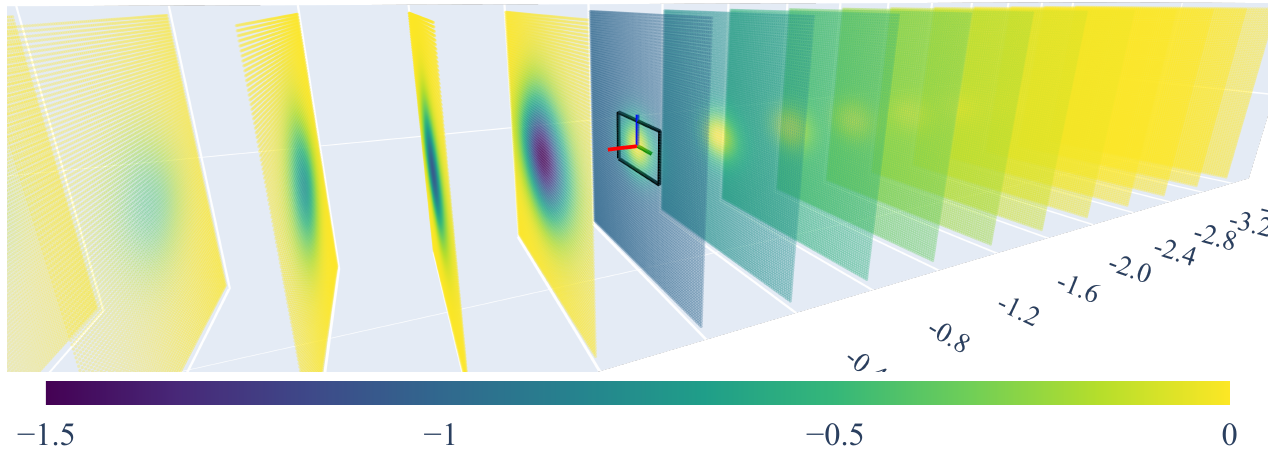}
    \caption{Illustration of part of the target waypoint's guidance reward field. The pass-through region (outlined using black lines) and the waypoint frame (RGB-$xyz$) are displayed at the center. The field spans to the entire $\mathbb{R}^3$.}
    \label{fig:r_guid}
\end{figure}

To calculate the guidance reward, we first need to transform the drone position from the world frame to the target waypoint frame. Let $\bm{p}_{\mathcal{G}}=[x\ y\ z]^{\mathsf{T}}$ denote drone position in the waypoint frame, the guidance reward is $r_{\text{guid}} = -f^2(x) \cdot g(x, y, z)$, with $f(x)=\max(1-\text{sgn}(x)x/k_0, 0)$ and $g(x,y,z)$ expanded to:

\begin{equation}
    g(x,y,z) = \begin{cases}
        k_1 \exp(-\frac{y^2+z^2}{2v}), & x > 0 \\
        1 - \exp(-\frac{y^2+z^2}{2v}), & x \leq 0
    \end{cases} \text{,}
\end{equation}

\noindent where $v$ is further expanded to:

\begin{equation}
    v = k_2 \left(1 + f^2(x)\right) \sqrt{\frac{z^2+y^2}{\left(z/h_{\text{wp}}\right)^2 + \left(y/w_{\text{wp}}\right)^2}} \text{,}
\end{equation}

\noindent if $y^2+z^2\neq0$. Otherwise $v=k_2 \left(1 + f^2(x)\right)$. Here $k_i$ for $i$ from 0 to 2 are scalar parameters, and $k_2$ is different for cases $x>0$ and $x\leq0$, i.e. for different sides of the waypoint. Figure \ref{fig:r_guid} illustrates the guidance reward field induced by the target waypoint. Our formulation adapts the original ``safety reward'' to rectangular waypoints and additionally penalizes the behavior of approaching the gate from the wrong side.

The waypoint passing reward $r_{\text{wp}}$ and the timeout reward $r_{\text{time}}$ are sparse and only become non-zero at specific steps: $r_{\text{wp}}$ turns to positive if the drone has just passed through a waypoint, and $r_{\text{time}}$ turns to negative if the drone has not crossed the final waypoint within a time limit.

Finally, we use the linear velocity reward $r_{\text{vel}}$ to encourage forward flight, which is beneficial for both making progress and obstacle avoidance with a limited camera field of view. It penalizes lateral and backwards velocity in the body frame $\bm{v}_{\mathcal{B}}=[v_x\, v_y\, v_z]^{\mathsf{T}}$ using negative parameters $k_3$ and $k_4$:

\begin{equation}
    r_{\text{vel}} = k_3 v_y^2 + k_4 \left(\min(v_x, 0)\right)^2 \text{.}
\end{equation}

\subsubsection{Observations}

We assume all observations are noise-free and deterministic. At time $t$, the observations include: the depth image $\bm{d}_t \in [0, 1]^{270\times480}$, drone states $\bm{s}_t \in [-1, 1]^{18}$, the last action $\bm{a}_{t-1} \in [-1, 1]^4$, and waypoint information of the next two target waypoints $\bm{w}_t \in [-1, 1]^{34}$.

The depth image is produced by a depth camera using the pinhole model. We set resolutions to 270$\times$480 and its horizontal FOV to 90 degrees. The transform from the ground-truth depth $\bm{d}_{\text{gt}}$ to the observed depth $\bm{d}$ is:

\begin{equation}
    \bm{d} = \min\left(\bm{d}_{\text{gt}}/d_{\text{max}},\bm{1}\right) \text{,}
\end{equation}

\noindent where $d_{\text{max}}$ denotes the maximum sensing range.

The drone states vector is defined as:

\begin{equation}
    \bm{s} = \begin{bmatrix}
        \frac{(\bm{p}_{\mathcal{W}} - \bm{p}_{\mathcal{W}_0})^{\mathsf{T}}}{p_{\text{max}}} & \bm{x}_{\mathcal{B}}^{\mathsf{T}} & \bm{y}_{\mathcal{B}}^{\mathsf{T}} & \bm{z}_{\mathcal{B}}^{\mathsf{T}} &
        \frac{\bm{v}_{\mathcal{W}}^{\mathsf{T}}}{v_{\text{max}}} &
        \frac{\bm{\omega}_{\mathcal{B}}^{\mathsf{T}}}{\omega_{\text{max}}}
    \end{bmatrix}^{\mathsf{T}} \text{,}
\end{equation}

\noindent where $\bm{p}_{\mathcal{W}_0}$ is the initial drone position; $\bm{x}_{\mathcal{B}}$, $\bm{y}_{\mathcal{B}}$, and $\bm{z}_{\mathcal{B}}$ are column vectors of the rotation matrix $\bm{R}_{\mathcal{WB}}$. Manually adjustable parameters for maximum sensing ranges for the position, linear velocity, and angular velocity are denoted by $p_{\text{max}}$, $v_{\text{max}}$, and $\omega_{\text{max}}$, respectively. This vector is further clamped to $[-1, 1]$ before returned.

We include information about two future waypoints based on the result of the gate observation experiment in \cite{adrdrl}: including information about two future gates can improve success rate and lap times. The information vector of one waypoint, indexed $i$, is defined as:

\begin{equation}
    \bm{w}_t^i = \begin{bmatrix}
        s_c & \min(\bm{l}^{\mathsf{T}}/{l_{\text{max}}},\bm{1}) & \bm{v}_{\text{corners}}^{\mathsf{T}}
    \end{bmatrix} ^ {\mathsf{T}} \text{,}
\end{equation}

\noindent with $s_c$ being the cosine similarity between vector $\bm{p}_{\text{wp}_i} - \bm{p}_{\mathcal{WB}}$ and the $x$-axis of waypoint $i$, $\bm{l}$ being the vector containing lengths of vectors from the origin of the drone body frame to 4 waypoint corners, $l_{\text{max}}$ being the maximum allowed value of these lengths, and $\bm{v}_{\text{corners}}$ denoting concatenated unit vectors from the drone to the corners. The dimension of $\bm{w}_t^i$ adds up to 17, so the dimension of $\bm{w}_t$, containing $\bm{w}_t^0$ and $\bm{w}_t^1$, is 34.

\subsubsection{Policy}

We use a neural network to represent the policy. Denoting the parameters of the neural network by $\bm{\theta}$, we can express the policy function as:

\begin{equation}
    \bm{a}_t = \pi_{\bm{\theta}}(\bm{d}_t, \bm{s}_t, \bm{w}_t, \bm{a}_{t-1}) \text{.}
\end{equation}

\noindent The neural network consists of an image encoder module that encodes an image $\bm{d}_t$ to a 64-dimensional latent vector $\bm{z}_t$ and a multi-layer perceptron (MLP) module that maps the concatenated 120-dimensional vector $[\bm{z}_t^{\mathsf{T}}\, \bm{s}_t^{\mathsf{T}}\, \bm{w}_t^{\mathsf{T}}\, \bm{a}_{t-1}^{\mathsf{T}}]^{\mathsf{T}}$ to action $\bm{a}_t$.

We employ a pre-trained Deep Collision Encoder (DCE) \cite{dce} as the image encoder and freeze its weights during training. The DCE is a residual network using convolutional layers at each block, and has fully connected layers that generate the mean and variance of the latent distribution, from which the latent vector $\bm{z}$ is sampled. This sampling process makes the policy stochastic.

The MLP module consists of two major parts. The first part is composed of 4 fully connected linear layers with Exponential Linear Unit (ELU) activation in between each layer, while the second part contains separate linear layers. The output of the first part is mapped to the mean and variance of the action distribution, in addition to a scalar, also known as the ``value'', by 3 separate linear layers. During training, the action is sampled from the distribution characterized by mean and variance, and during evaluation and deployment, the action directly uses the mean.

\subsection{Policy Training}

Given that our policy is represented by a neural network, our goal boils down to finding the optimal parameters $\bm{\theta}$ that maximizes the expectation of accumulated rewards in Equation (\ref{eq:pomdp_goal}). We use the Proximal Policy Optimization (PPO) \cite{ppo} reinforcement learning algorithm for this purpose. Additionally, we explore the technique of domain randomization to enable generalization to unseen environments, hoping that with enough variability encountered during training, an unseen environment would appear as just another variation, where the policy has required knowledge to finish the track. To achieve this, we have designed a waypoint generator, a random obstacle manager, and multiple training strategies.

\subsubsection{Waypoint Generator and Obstacle Manager}

\begin{figure}[t]
    \centering
    \includegraphics[width=\linewidth]{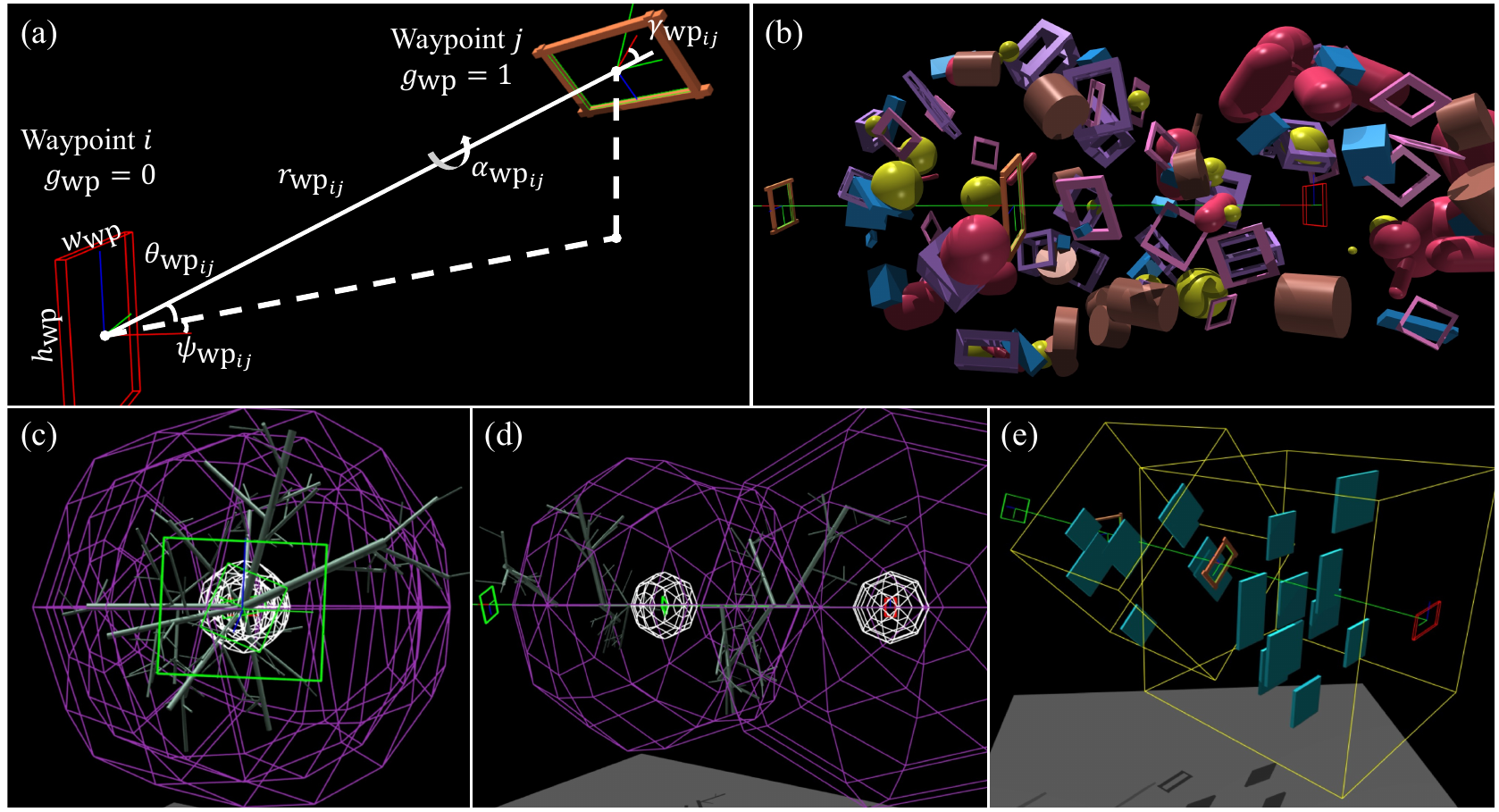}
    \caption{Illustration of parameters describing relative waypoint poses (a) and obstacles managed by the obstacle manager (b)-(e). Sub-figure (b) shows orbital obstacles, (c) and (d) show tree-like obstacles from different views, and (e) shows wall-like obstacles between waypoints.}
    \label{fig:wp_obs}
\end{figure}

We consider a racing track the combination of waypoints and obstacles for the obstacle-aware drone racing task. We use the waypoint generator to generate random waypoints and use the obstacle manager to put obstacles at places that effectively block the flight paths between waypoints.

We use 5 values $(\psi_{\text{wp}_{ij}}, \theta_{\text{wp}_{ij}}, r_{\text{wp}_{ij}}, \alpha_{\text{wp}_{ij}}, \gamma_{\text{wp}_{ij}})$ to parameterize relative pose between waypoint $i$ and $j$, as shown in Figure \ref{fig:wp_obs}(a). Given the pose of waypoint $i$ as position vector and rotation matrix $(\bm{p}_{\text{wp}_i}, \bm{R}_{\text{wp}_i})$, the pose of waypoint $j$ can be calculated using:

\begin{equation}
\label{eq:wp_tf}
    \begin{split}
        \bm{p}_{\text{wp}_j} &= r_{\text{wp}_{ij}} \bm{R}_y(\theta_{\text{wp}_{ij}}) \bm{R}_z(\psi_{\text{wp}_{ij}}) \bm{R}_{\text{wp}_i}\begin{bmatrix}
            1 & 0 & 0
        \end{bmatrix} ^{\mathsf{T}} + \bm{p}_{\text{wp}_i} \\
        \bm{R}_{\text{wp}_j} &= \bm{R}_y(\gamma_{\text{wp}_{ij}}) \bm{R}_x(\alpha_{\text{wp}_{ij}}) \bm{R}_y(\theta_{\text{wp}_{ij}}) \bm{R}_z(\psi_{\text{wp}_{ij}}) \bm{R}_{\text{wp}_i}
    \end{split} \text{.}
\end{equation}

\noindent Although there are only 5 degrees of freedom, this parameterization allows for enough room for randomization and intuitive adjustments of the track's difficulties.

Waypoints are generated procedurally. Firstly, the Initial waypoint's roll, pitch, and yaw angles are sampled uniformly within defined bounds, and the position is set to an arbitrary value. Secondly, for subsequent waypoints, we sample the relative pose parameters uniformly within defined bounds and calculate their poses till the final waypoint using Equation (\ref{eq:wp_tf}). Thirdly, parameters $(w_{\text{wp}}, h_{\text{wp}}, g_{\text{wp}})$ are also sampled uniformly within defined ranges for all waypoints. Lastly, we offset all waypoints' positions to fit the track within environment boundaries.

We observe that uniformly distributing obstacles in $\mathbb{R}^3$, as seen in \cite{ntnu-rlflight}, is not suitable for significantly larger environments. Uniformly distributing obstacles requires an excessive number of obstacles in the environments, which increases computational overhead and hurts simulation performance. Our obstacle manager allows for challenging the agent on obstacle avoidance using a small number of obstacles, by strategically sampling obstacle poses based on the generated waypoints. The manager supports uniformly distributing tree-like obstacles along line segments connecting waypoint centers, placing wall-like cuboids between waypoints, and finally making obstacles of various shapes orbit waypoints. Managed obstacles are illustrated in Figure \ref{fig:wp_obs}(b) to \ref{fig:wp_obs}(e). By anchoring obstacles to the racing track, we achieve efficient obstacle management. Furthermore, the difficulty level can controlled by specifying the number of obstacles in each group and parameters defining the shapes of the obstacles.

\subsubsection{Training on Track Segments}

Since full-length tracks are the combination of segments of shorter lengths, we believe that training on short track segments will allow generalization to full-length tracks, while reducing computational overhead for the waypoint generator, obstacle manager, and the physics engine. Plus, with shorter track lengths, we can fit full episodes into shorter rollout horizons, which increases policy update frequency and potentially reduces the wall-clock time required for the policy to converge. We generate waypoints and manage obstacles for short track segments containing only 4 waypoints. The task is to fly from the initial position near waypoint 0, pass through waypoint 1, and finally finish the episode by passing through waypoint 2. Waypoint 3 is generated to keep the dimensions of the waypoint information vector $\bm{w}_t$ consistent.

\subsubsection{Environment Randomization}

Combining the waypoint generator and obstacle manager, we can create an infinite amount of random environments for training. Aiming to provide the agent with diverse experience, our implementation of the waypoint generator and the obstacle manager supports vectorized environments, that is, for a single round of data collection (rollout), multiple different environments are randomly created using these tools. By incorporating experience collected in different environments into a single rollout dataset, we avoid overfitting the policy to a single environment from the ground up. For a small number of parallel environments, e.g. a few hundred, creating only a single set of random environments and using them for all rollouts is not enough, as this makes the policy learn an average strategy that maximizes the mean total reward for this specific set of environments. To solve this problem, we generate a new set of random environments for every single rollout, which further diversifies the total experience dataset.

\subsubsection{Random Agent Initialization}

The camera transform in the drone body frame and drone states are randomly initialized upon agent resets to make the policy more robust. This strategy can also be seen as an application of domain randomization. Specifically, we randomize the camera position in the $yz$-plane of the drone body frame, and the camera tilt angle. For drone states, we initialize the position within the obstacle-free zone of the initial waypoint to avoid spawning the drone into obstacles, other states such as linear and angular velocities, as well as the initial actions, are uniformly sampled within defined ranges. The initial attitude is either the same as the initial waypoint's attitude or the resulting attitude of firstly aligning the body $x$-axis with vector $\bm{p}_{\text{wp}_1}-\bm{p}_{\text{wp}_0}$, and secondly rotating about the vector for a random angle.

\subsection{Implementation Details}

We implement vectorized environments based on the Isaac Gym Simulator \cite{isaacgym}, which supports parallel physics simulation on GPUs and offers relatively high image rendering speed. To work with Isaac Gym, our code is highly optimized using PyTorch-based vectorized operations. The physics simulation frequency and angular velocity control frequency are set to 250 Hz, while the camera rendering frequency and policy control frequency are set to 25 Hz, that is, one environment step corresponds to 10 closed-loop physics steps. With this setting, we achieve about 3,000 total environment steps per second with camera sensors enabled. This speed is recorded on a consumer-grade desktop PC equipped with an Intel i9-13900K CPU and an Nvidia RTX 4090 GPU running the Ubuntu 22.04 operating system.

\begin{figure}[b]
    \centering
    \includegraphics[width=\linewidth]{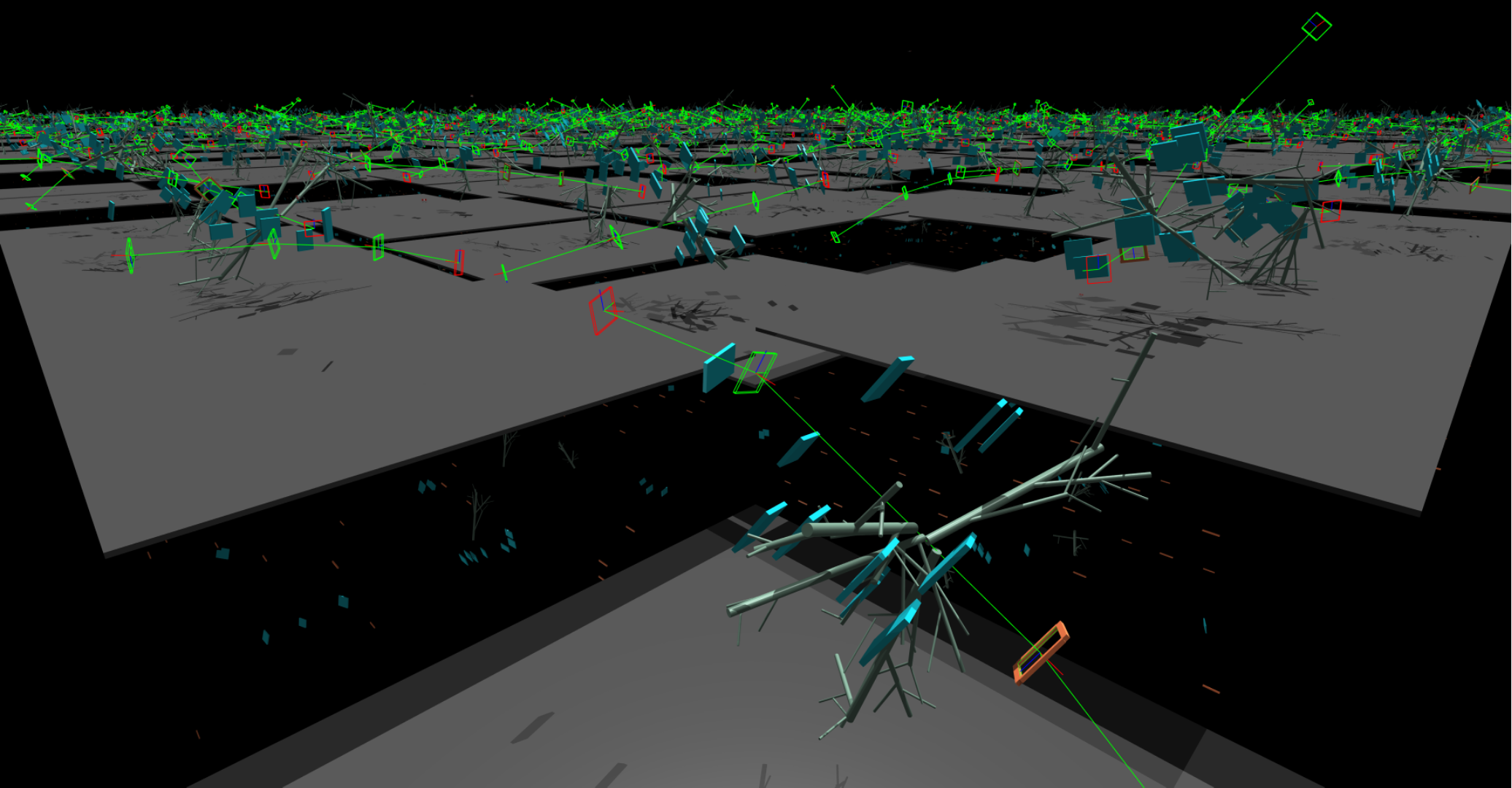}
    \caption{Illustration of parallel environments for training. Environments are tiled up in Isaac Gym but are independent and asynchronous. Debug views of the waypoints are enabled here for visualization purposes, but are disabled during actual training.}
    \label{fig:train_env}
\end{figure}

\begin{figure}[t]
    \centering
    \includegraphics[width=\linewidth]{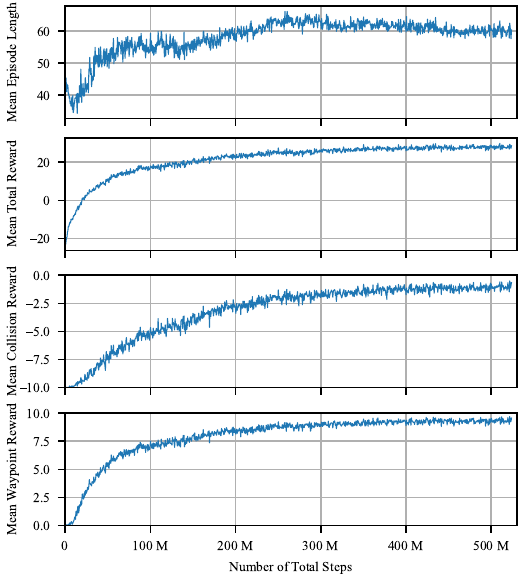}
    \caption{Logged mean episode length in steps, mean total reward, mean collision reward, and mean waypoint reward throughout training.}
    \label{fig:train_log}
\end{figure}

Although the implemented waypoint generator and obstacle manager are capable of generating complex racing tracks with a large number of obstacles, for now, we train the policy in simpler environments. To ensure enough experience diversity within data collected through one rollout, we run 512 random environments in parallel. Every environment includes 4 tree-like obstacles taken from Aerial Gym \cite{aerialgym}, and 12 wall-like obstacles with sizes randomly specified from $(0.2, 1.5, 1.5)$ to $(0.2, 2.5, 2.5)$ in meters. For the waypoint generator, the range of waypoint sizes is set to $[1.4, 2.0]$; the initial waypoints' roll and pitch are restricted within range $[-0.2, 0.2]$, but yaw can be an arbitrary value; bounds of the relative waypoint pose parameters $(\psi_{\text{wp}_{ij}}, \theta_{\text{wp}_{ij}}, r_{\text{wp}_{ij}}, \alpha_{\text{wp}_{ij}}, \gamma_{\text{wp}_{ij}})$ are $(-0.3, -0.3, 6, 0, 0)$ and $(0.3, 0.3, 18, 3.14, 0.2)$. Figure \ref{fig:train_env} shows the appearances of such environments.

We code the training loop based on the actor-critic PPO agent in RL-Games \cite{rl-games}. Our domain randomization happens during environment resets, so we modify the original training loop to include calling environment reset before running rollout in every training iteration. In total, we train the policy for 1,000 iterations, which corresponds to collecting experiences in 512,000 different environments for about 520 million environment steps. It takes about 50 wall-clock hours to finish all iterations. Figure \ref{fig:train_log} shows the mean episode length, total reward, and the collision and waypoint reward terms throughout training. With the collision reward set to -10 at collision, and the waypoint reward set to 5 at waypoint passing, the corresponding reward curves suggest achieving around 10\% crash rate and 90\% success rate at the end of training.

\section{Experiments}

\subsection{Demonstrating Generalizable Obstacle-Aware Racing}

\begin{figure*}[t]
    \centering
    \includegraphics[width=\textwidth]{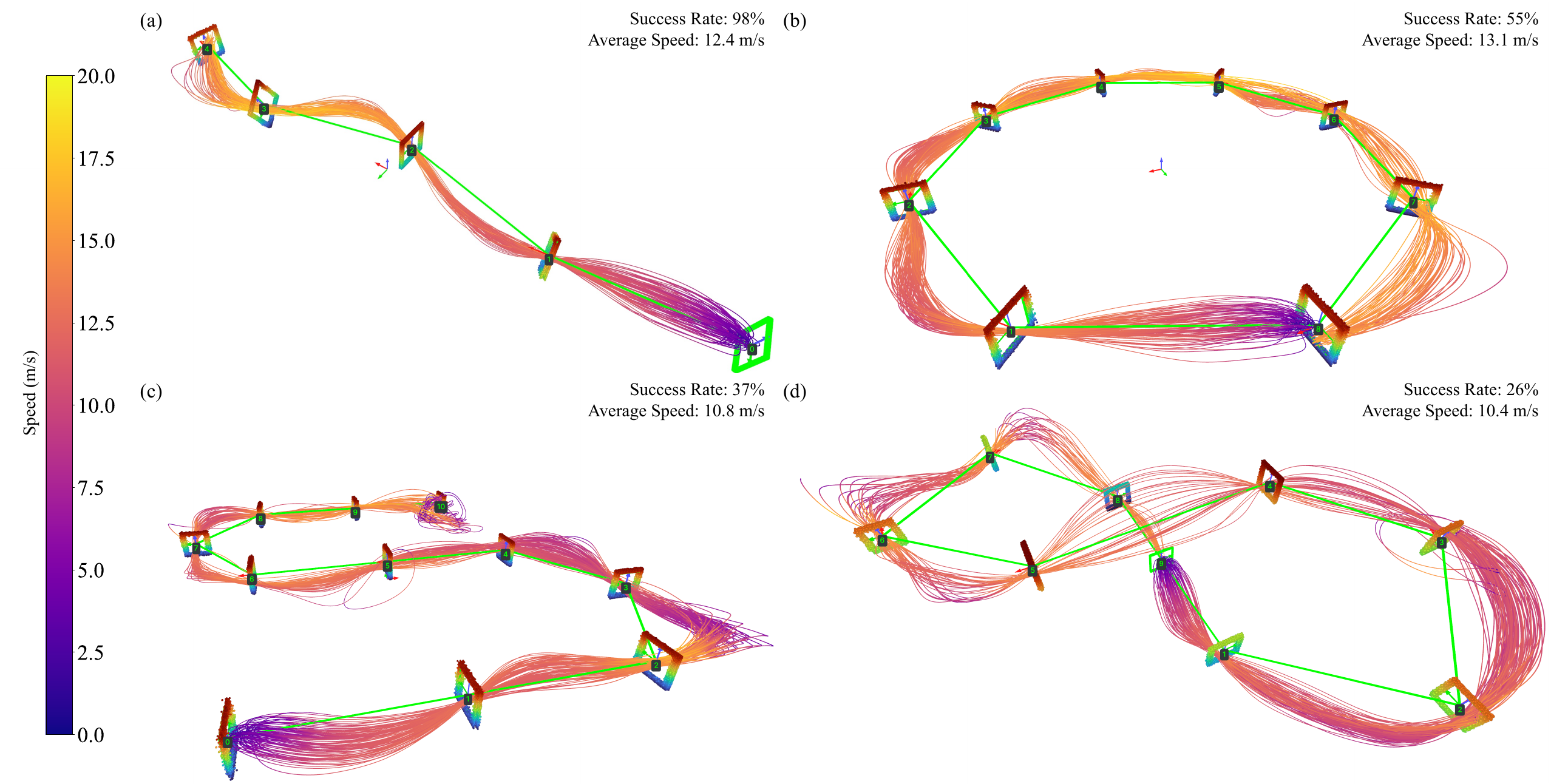}
    \caption{Illustration of rollout trajectories of the trained policy on multiple different racing tracks without obstacles: (a) ``Kebab'', (b) ``Circle'', (c) ``Turns'', and (d) ``Wavy Eight''. Physical gate bars are represented as points colored according to their $z$ positions in the world frame.}
    \label{fig:test_no_obst}
\end{figure*}

\begin{figure}[t]
    \centering
    \includegraphics[width=\linewidth]{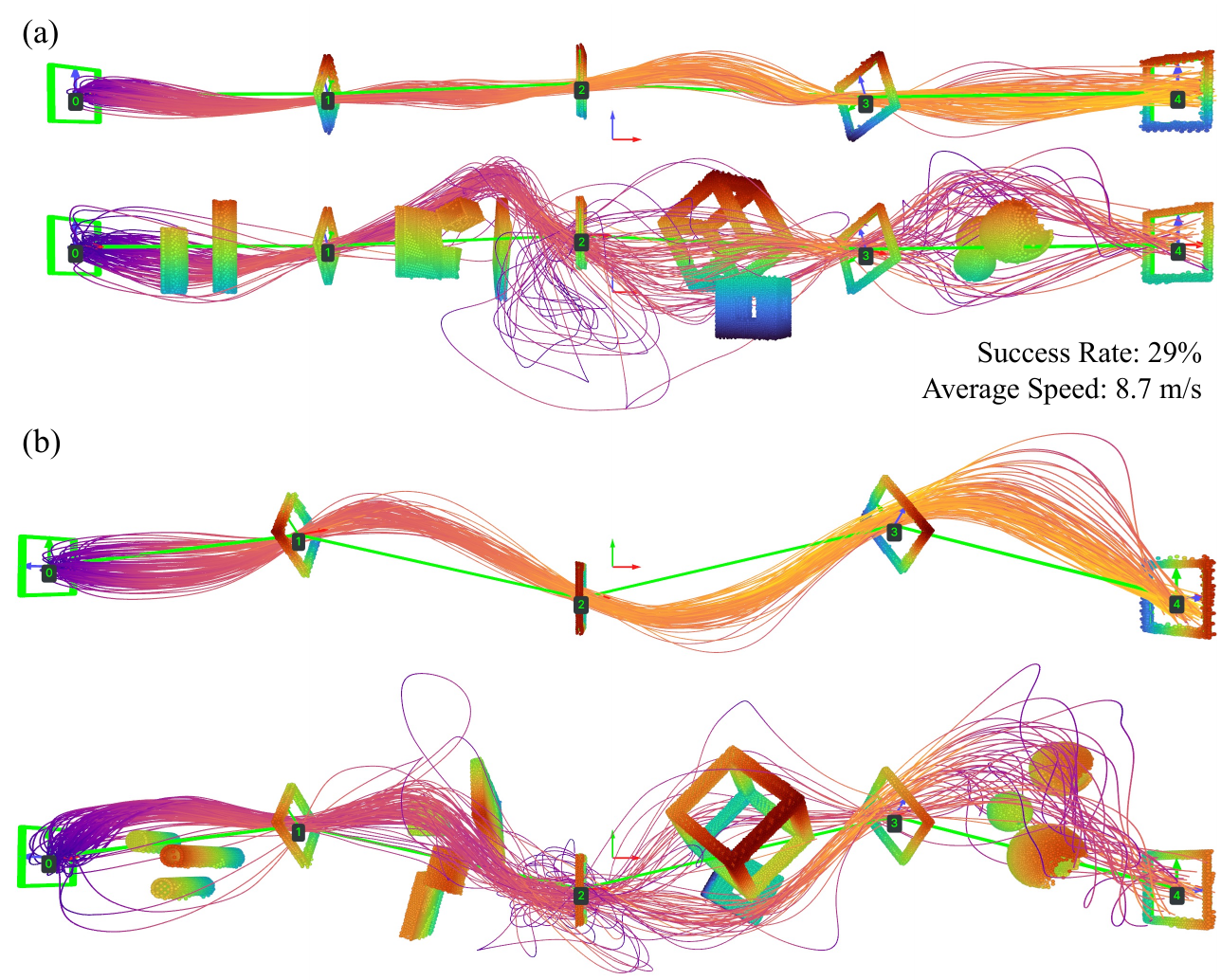}
    \caption{Side views (a) and top-down views (b) showing distorted trajectories due to the presence of obstacles.}
    \label{fig:test_kebab}
\end{figure}

\begin{figure}[t]
    \centering
    \includegraphics[width=\linewidth]{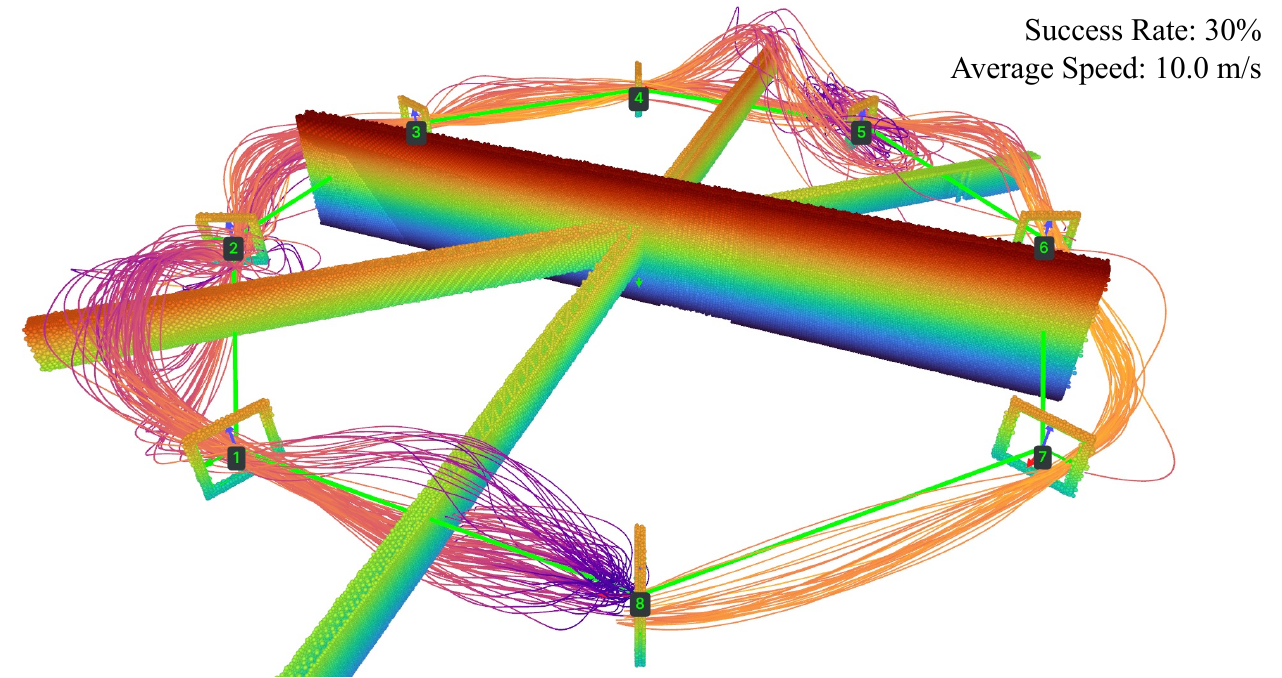}
    \caption{Trajectories executed on the ``Circle'' track with additional obstacles.}
    \label{fig:test_circle}
\end{figure}

We first demonstrate the policy's ability to generalize to unseen waypoint placement and the number of waypoints. Four full-length tracks are designed for this experiment: ``Kebab'', ``Circle'', ``Turns'', and ``Wavy Eight''. The ``Kebab'' features 5 waypoints roughly in a row, representing scenarios encountered during training. The ``Circle'' consists of 9 waypoints uniformly distributed on a circle with a radius of 15 meters. The ``Turns'' has 11 waypoints positioned on a big letter ``S''. Then in the ``Wavy Eight'', waypoints form a figure of eight, with their $z$ coordinates different. Since the drone's position is part of the observation, all waypoints in the test tracks are positioned within the same environment boundaries as those used for training.

Drones are randomly initialized as in training, but we set larger ranges for initial attitude, body-frame velocities, and commands, to further ``stress'' the policy. Under this setup, we roll out 100 episodes for each track, log the trajectories, and calculate the success rate and the average linear speed. Results are shown in Figure \ref{fig:test_no_obst}. On track ``Kebab'', the most similar to the training scenarios, the policy achieves the highest success rate. As the track gets more twists and includes more sharp turns that are not present in the training set, the success rate drops. Despite the performance drop, this experiment confirms that the policy generalizes to full-length racing tracks and completely unseen relative waypoint poses.

To verify the generalizable obstacle avoidance ability of the policy, we deliberately place obstacles of various shapes, including ones not present in training, on track ``Kebab'' and ``Circle'' to block trajectories that might have been executed if there were no obstacles. If the policy is capable of obstacle avoidance, the resulting trajectories would be distorted. Furthermore, due to randomness in the initial drone states, we also anticipate observing non-homotopic trajectories or passing around obstacles on different sides. The results shown in Figures \ref{fig:test_kebab} and \ref{fig:test_circle} confirm that our policy has indeed learned the generalizable ability to avoid obstacles. However in several parts of these designed tracks, the obstacle configurations have deviated too far from those in training environments, so the success rates are low.

\subsection{Benchmarking Policy by Varying Scene Complexity}

\begin{figure*}[t]
    \centering
    \includegraphics[width=\textwidth]{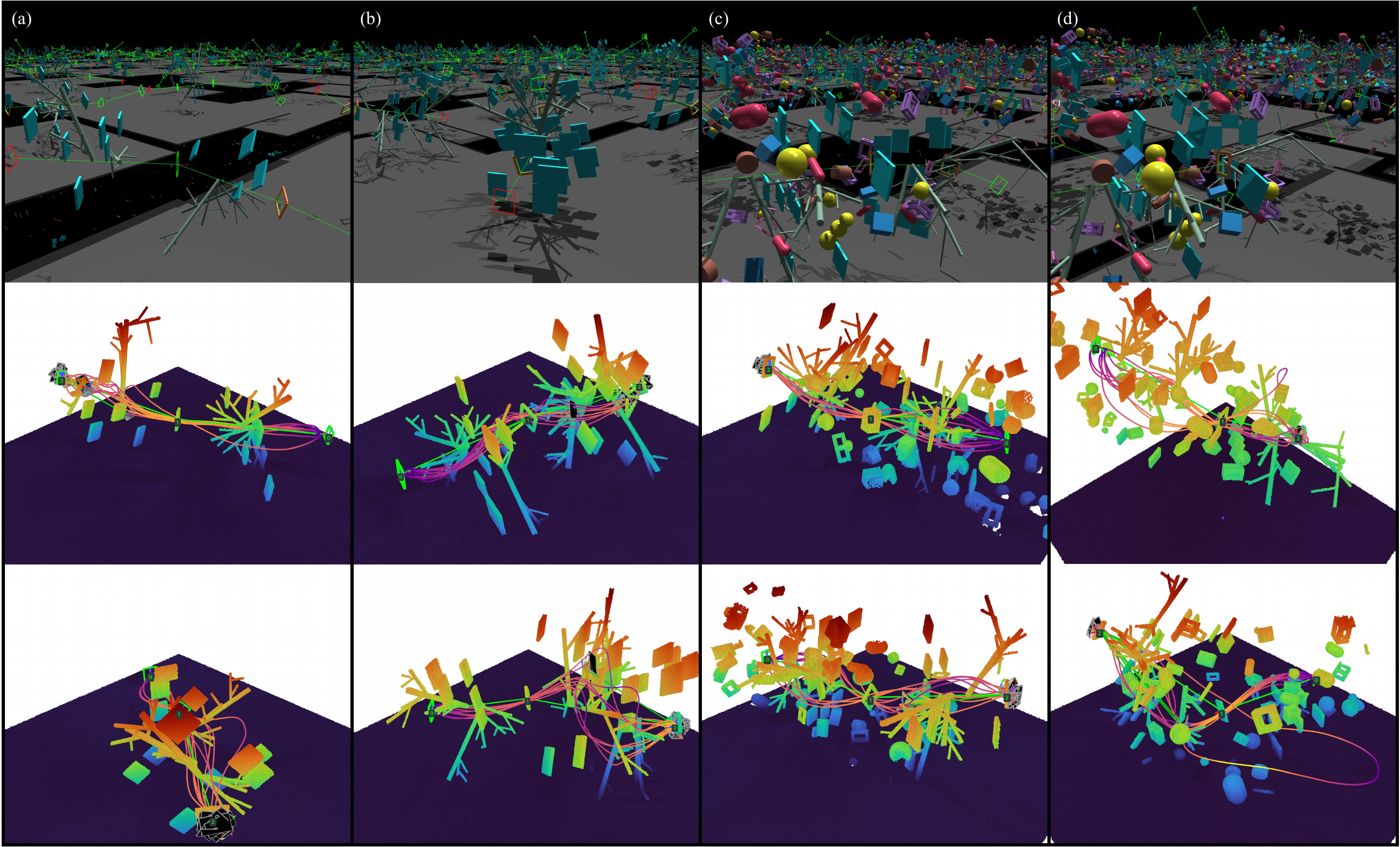}
    \caption{Illustration of random tracks of different complexities and executed trajectories on two randomly selected tracks. Sub-figures (a) to (b) correspond to difficulty levels from 1 to 4.}
    \label{fig:test_rand}
\end{figure*}

\begin{figure}[t]
    \centering
    \includegraphics[width=\linewidth]{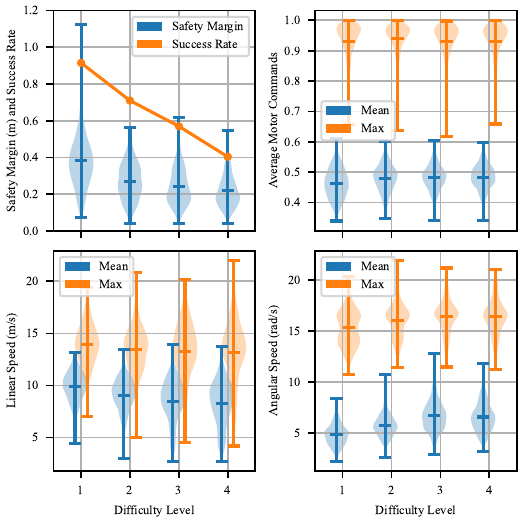}
    \caption{Policy performance metrics across different difficulty levels. The safety margin of an episode is the minimum distance between the trajectory and the obstacle.}
    \label{fig:policy_metrics}
\end{figure}

To further evaluate the capability of our policy, we assess its performance on a series of randomized tracks with varying levels of complexity. Level 1 represents the difficulty level of training environments: in each environment there are 12 wall-like obstacles and 4 tree-like obstacles, relative waypoint pose parameters $(\psi_{\text{wp}_{ij}}, \theta_{\text{wp}_{ij}}, r_{\text{wp}_{ij}}, \alpha_{\text{wp}_{ij}}, \gamma_{\text{wp}_{ij}})$ are set between $(-0.3, -0.3, 6, 0, 0)$ and $(0.3, 0.3, 18, 3.14, 0.2)$. Level 2 doubles the amount of obstacles and leaves waypoint parameters unchanged. Level 3 includes 60 additional obstacles orbiting waypoint 0 and waypoint 1. Finally, level 4 sets the bounds of the waypoint parameters to $(-1, -0.4, 6, 0, 0)$ and $(1, 0.4, 18, 3.14, 0.3)$ on top of other settings in level 3.

We generate 100 random tracks per difficulty level and roll out 10 episodes per track, resulting in a total of 1000 episodes per level. Screenshots of environments in Isaac Gym, sample environments, and resulting trajectories are illustrated in Figure \ref{fig:test_rand}. For each trajectory, we log its termination mode, safety margin, mean and maximum values of average commands of all motors, linear speed, and angular speed. Then we can calculate the success rates and plot the data distributions of other metrics over all trajectories for all difficulty levels, as shown in Figure \ref{fig:policy_metrics}.

The success rate starts at 0.9 for level 1, consistent with the training results shown in Figure \ref{fig:train_log}, but decreases as track difficulty increases, reaching around 0.4 by level 4. A similar downward trend is observed for the safety margin, indicating that the drone comes closer to obstacles on harder tracks. In terms of control effort, the maximum values of motor commands remain consistent across difficulty levels, suggesting that the policy pushes the drone to its control limits whenever possible. The mean values of commands show a clear upward trend, implying that navigating more complex tracks requires increasingly aggressive control inputs. This is also reflected in the angular speed, where both mean and maximum values increase with difficulty, as the drone must rotate more quickly to handle tighter turns and avoid obstacles. Finally, both the mean and maximum values of linear speed show a slight decrease as difficulty increases, due to the drone slowing down in response to more complex track layouts and more cluttered spaces.

These results show that the policy generalizes to unseen tracks and achieves a high success rate in similar-to-training environments. As the test set deviates from the training set, the policy can adapt to increased difficulty with higher control effort and more careful velocity management and still maintain a certain level of success rate.

\subsection{Towards Robust Obstacle-Aware Racing Policies}

\begin{figure}[t]
    \centering
    \includegraphics[width=\linewidth]{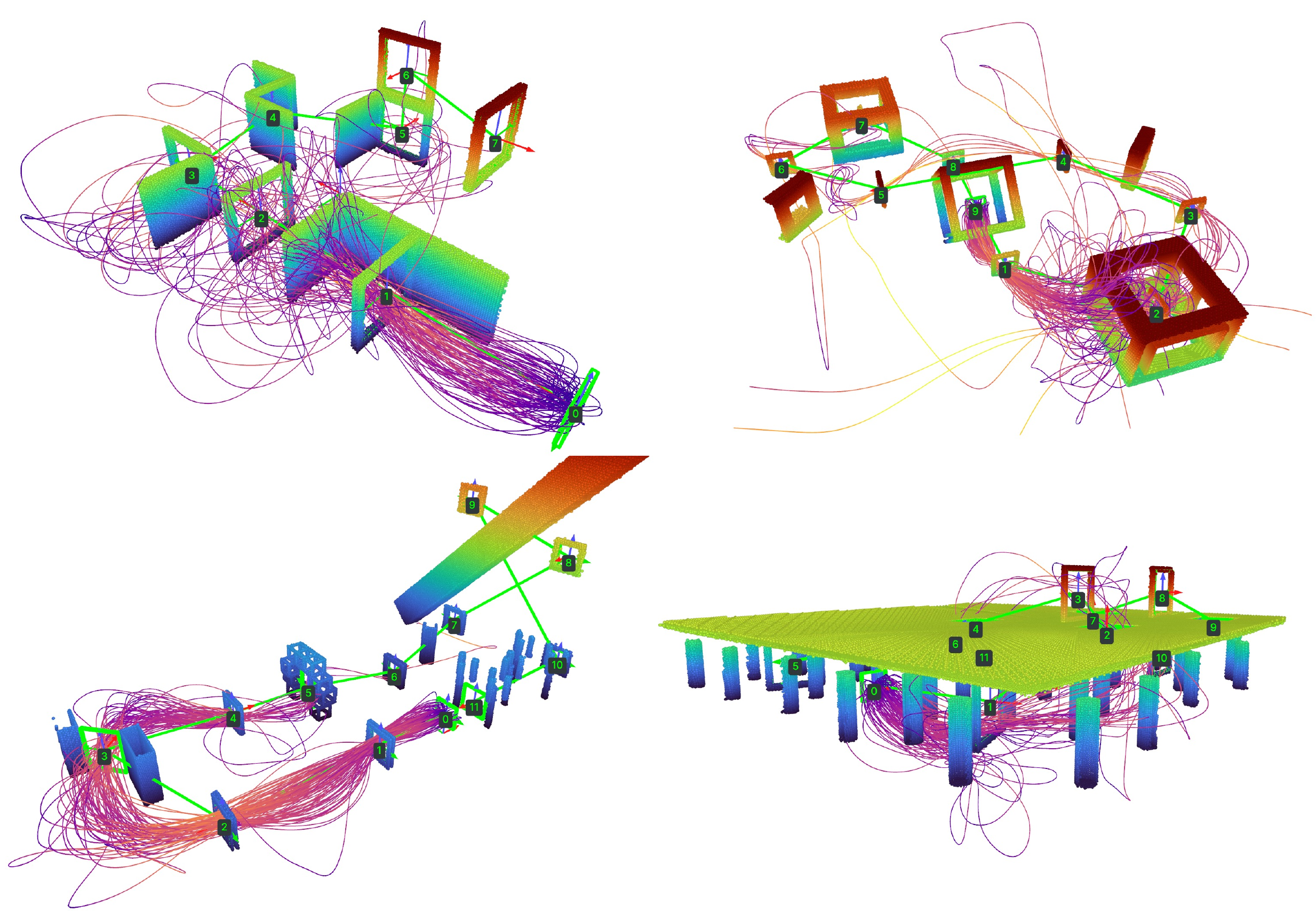}
    \caption{Illustration of rollout trajectories on hard racing tracks where distances between waypoints are much shorter, requiring sharper turns. Most trajectories are unsuccessful.}
    \label{fig:test_hard}
\end{figure}

We additionally evaluate the policy on four hard tracks, all characterized by shorter waypoint distances, and a higher density of obstacles. These characteristics require the drone to do sharper turns. As a result, the policy struggles to navigate, with most trajectories being unsuccessful, lowering success rates to below 0.01, as shown in Figure \ref{fig:test_hard}. This performance decline suggests that the current policy, trained in simple environments, lacks the robustness required to navigate more complex and constrained tracks. This limitation could be due to over-simplified environments in the training set. The training tracks have fewer obstacles, and obstacles are all distributed in a simple way, which may make the policy overfit to this specific track design. As a result, the policy struggles in new, more challenging scenarios that require advanced obstacle avoidance and tighter maneuvering.

To improve robustness, the core problem would be how to randomly create a set of training environments that better represent the complexity of real-world tracks. Once this problem is answered, we can train the policy with domain randomization to finally obtain a robust policy for obstacle-aware drone racing.

\section{Conclusion}

This work presents an approach for training a generalizable obstacle-aware drone racing policy using domain randomization and deep reinforcement learning. The policy is trained on randomized short track segments, and evaluated on both hand-crafted, full-length tracks and randomized short segments across varying difficulty levels. Experiment results demonstrate that the policy generalizes well to unseen tracks and adapts to increased difficulty levels, achieving high success rates in environments closely resembling the training set. However, the policy encounters challenges in more complex, cluttered environments, where obstacle density and tighter waypoint spacing result in significant performance drops of the policy. This highlights the importance of further research into the method for generating more diverse and challenging training environments. Future work may also explore using advanced training strategies such as curriculum learning or adaptive difficulty scaling to better prepare policies for real-world drone racing challenges.

\bibliographystyle{IEEEtran}
\bibliography{references}

\end{document}